\documentclass[10pt,twocolumn,letterpaper]{article}

\usepackage{cvpr}
 
\usepackage[margin=0.93in]{geometry}
\usepackage{times}
\usepackage{epsfig}
\usepackage{graphicx}
\usepackage{grffile}
\usepackage{amsmath}
\usepackage{amssymb}
\usepackage{color}
\usepackage{float}
\usepackage{authblk}
\usepackage[symbol*]{footmisc}

\DefineFNsymbolsTM{otherfnsymbols}{%
  \textexclamdown \wr
  \textsection   \mathsection
  \textdagger    \dagger
  \textdaggerdbl \ddagger
  \textasteriskcentered *
  \textbardbl    \|%
  \textparagraph \mathparagraph
}%

\setfnsymbol{otherfnsymbols}

\usepackage[pagebackref=true,breaklinks=true,letterpaper=true,colorlinks,bookmarks=false]{hyperref}

\cvprfinalcopy 

\def\cvprPaperID{3092} 

\ifcvprfinal\fi

\definecolor{rowblue}{RGB}{220,230,240}
\definecolor{myred}{RGB}{240,20,10}
\definecolor{mygreen}{RGB}{0,150,100}
\definecolor{myblue}{RGB}{20,10,240}

\usepackage{amsmath}
\DeclareMathOperator*{\argmin}{arg\,min}

\begin{document}


\providecommand{\shortcite}[1]{\cite{#1}}

\title{High-Resolution Image Inpainting using Multi-Scale Neural Patch Synthesis}

\author[1]{Chao Yang\thanks{chaoy@usc.edu}}
\author[2]{Xin Lu\thanks{xinl@adobe.com}}
\author[2]{Zhe Lin\thanks{zlin@adobe.com}}
\author[2]{Eli Shechtman\thanks{elishe@adobe.com}}
\author[2]{Oliver Wang\thanks{owang@adobe.com}}
\author[1,3,4]{Hao Li\thanks{hao@hao-li.com}}
\affil[1]{University of Southern California}
\affil[2]{Adobe Research}
\affil[3]{Pinscreen}
\affil[4]{USC Institute for Creative Technologies}

\maketitle
\thispagestyle{empty}

\begin{abstract}
%
Recent advances in deep learning have shown exciting promise in filling large holes in natural images with semantically plausible and context aware details, impacting fundamental image manipulation tasks such as object removal. While these learning-based methods are significantly more effective in capturing high-level features than prior techniques, they can only handle very low-resolution inputs due to memory limitations and difficulty in training. Even for slightly larger images, the inpainted regions would appear blurry and unpleasant boundaries become visible. We propose a multi-scale neural patch synthesis approach based on joint optimization of image content and texture constraints, which not only preserves contextual structures but also produces high-frequency details by matching and adapting patches with the most similar mid-layer feature correlations of a deep classification network. We evaluate our method on the ImageNet and Paris Streetview datasets and achieved state-of-the-art inpainting accuracy. We show our approach produces sharper and more coherent results than prior methods, especially for high-resolution images. 

\end{abstract}

\section{Introduction}

\begin{figure}[!h]
\centering
\small
\begin{tabular}{cc}

  \includegraphics[width=.2\textwidth]{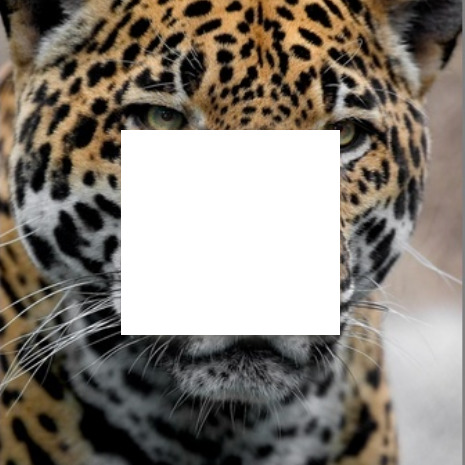}&
  \includegraphics[width=.2\textwidth]{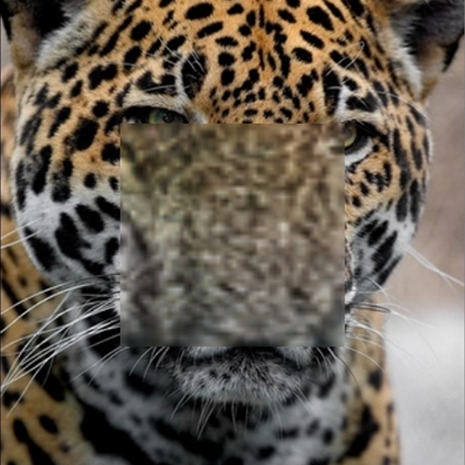}\\
  (a) Input Image & (b) Context Encoder \\
  \includegraphics[width=.2\textwidth]{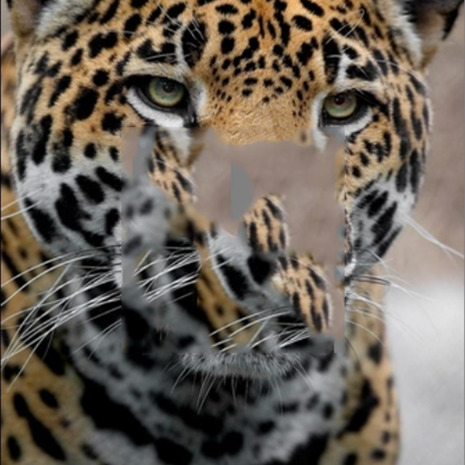}&
  \includegraphics[width=.2\textwidth]{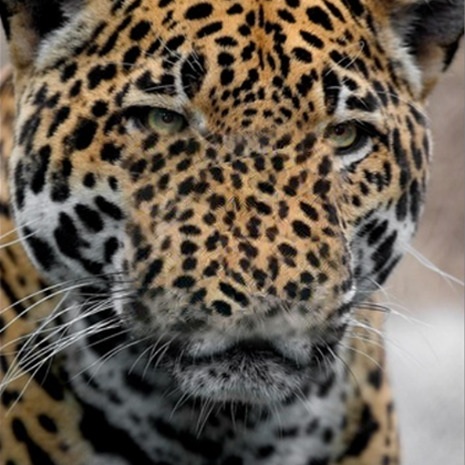} \\
(c) PatchMatch & (d) Our Result \\
\end{tabular}
\caption{Qualitative illustration of the task. Given an image ($512\times 512$) with a missing hole ($256\times 256$) (a), our algorithm can synthesize sharper and more coherent hole content (d) comparing with Context Encoder~\cite{philip} (b) and Content-Aware Fill using PatchMatch~\cite{patchmatch2} (c). }
\label{fig:teaser}
\end{figure}

Before sharing a photo, users may want to make modifications such as erasing distracting scene elements, adjusting object positions in an image for better composition, or recovering the image content in occluded image areas.
These, and many other editing operations, require automated hole-filling (image completion), which has been an active research topic in the computer vision and graphics communities for the past few decades.
Due to its inherent ambiguity and the complexity of natural images, general hole-filling remains challenging.

Existing methods that address the hole-filling problem fall into two groups.
The first group of approaches relies on texture synthesis techniques, which fills in the hole by extending textures from surrounding regions~\cite{efros99,efros01,kwatra03,kwatra05,criminisi03,drori03,wexler04,wilczkowiak05,komodakis06,komodakis07,patchmatch}.
A common idea in these techniques is to synthesize the content of the hole region in a coarse to fine manner, using patches of similar textures. In \cite{drori03,wilczkowiak05}, multiple scales and orientations are introduced to find better matching patches.
Barnes et al.~\shortcite{patchmatch} proposed PatchMatch as a fast approximate nearest neighbor patch search algorithm. Although such methods are good at propagating high-frequency texture details, they do not capture the semantics or global structure of the image. The second group of approaches hallucinates missing image regions in a data-driven fashion, leveraging large external databases.
These approaches assume that regions surrounded by similar context likely possess similar content~\cite{imgcompletion07}.
This approach is very effective when it finds an example image with sufficient visual similarity to the query but could fail when the query image is not well represented in the database. Additionally, such methods require access to the external database, which greatly restricts possible application scenarios.

More recently, deep neural network is introduced for texture synthesis and image stylization~\cite{gatys,gatysnips,chuanli,compositiontransfer,texturenetwork,percepturalloss}. In particular, Phatak et al.~\shortcite{philip} trained an encoder-decoder CNN (Context Encoder) with combined $\ell_2$ and adversarial loss~\cite{gan} to directly predict missing image regions.
This work is able to predict plausible image structures, and is very fast to evaluate, as the hole region is predicted in a single forward pass. Although the results are encouraging, the inpainting results of this method sometimes lack fine texture details, which creates visible artifacts around the border of the hole. This method is also unable to handle high-resolution images due to the difficulty of training regarding adversarial loss when the input is large.

In a recent work, Li and Wand~\shortcite{chuanli} showed that impressive image stylization results can be achieved by optimizing for an image whose neural response at mid-layer is similar to that of a content image, and whose local responses at a low convolutional layers resemble local responses from a style image. 
Those local responses were represented by small (typically $3 \times 3$) {\em neural patches}. This method proves able to transfer high-frequency details from the style image to the content image, hence suitable for realistic transfer tasks (e.g., transfer of the look of faces or cars). Nevertheless, transferring of more artistic styles are better addressed by using gram matrices of neural responses~\cite{gatys}.

To overcome the limitations of aforementioned methods, we propose a hybrid optimization approach that leverages the structured prediction power of encoder-decoder CNN and the power of neural patches to synthesize realistic, high-frequency details. Similar to the style transfer task, our approach treats the encoder-decoder prediction as the global content constraint, and the local neural patch similarity between the hole and the known region as the texture constraint. 

More specifically, the content constraint can be constructed by training a global content prediction network similar to Context Encoder, and the texture constraint can be modeled with the image content surrounding the hole, using the patch response of the intermediate layers using the pre-trained classification network. The two constraints can be optimized using backpropagation with limited-memory BFGS. In order to further handle high-resolution images with large holes, we propose a multi-scale neural patch synthesis approach. For simplicity of formulation, we assume the test image is always cropped to $512 \times 512$ with a $256 \times 256$ hole in the center. We then create a three-level pyramid with step-size two, downsizing the image by half at each level. It renders the lowest resolution of a $128 \times 128$ image with a $64 \times 64$ hole. We then perform the hole filling task in a coarse-to-fine manner. Initialized with the output of content prediction network at the lowest level, at each scale (1) we perform the joint optimization to update the hole, (2) upsample to initialize the joint optimization and set content constraint for the next scale. We then repeat this until the joint optimization is finished at the highest resolution (Sec.~\ref{sec:approach}). 

We show experimentally that the proposed multi-scale neural patch synthesis approach can generate more realistic and coherent results preserving both the structure and texture details. We evaluate the proposed method quantitatively and qualitatively on two public datasets and demonstrate its effectiveness over various baselines and existing techniques as shown in Fig. \ref{fig:teaser} (Sec.~\ref{sec:exp}).

The main contributions of this paper are summarized as follows:
\begin{itemize}
\item We propose a joint optimization framework that can hallucinates missing image regions by modeling a global content constraint and local texture constraint with convolutional neural networks.
\item We further introduce a multi-scale neural patch synthesis algorithm for high-resolution image inpainting based on the joint optimization framework.
\item We show that features extracted from middle layers of the neural network could be used to synthesize realistic image contents and textures, in addition to previous works that use them to transfer artistic styles.
\end{itemize}


\section{Related Work}

\begin{figure*}[t]
	\centering
	\includegraphics[width=.9\linewidth]{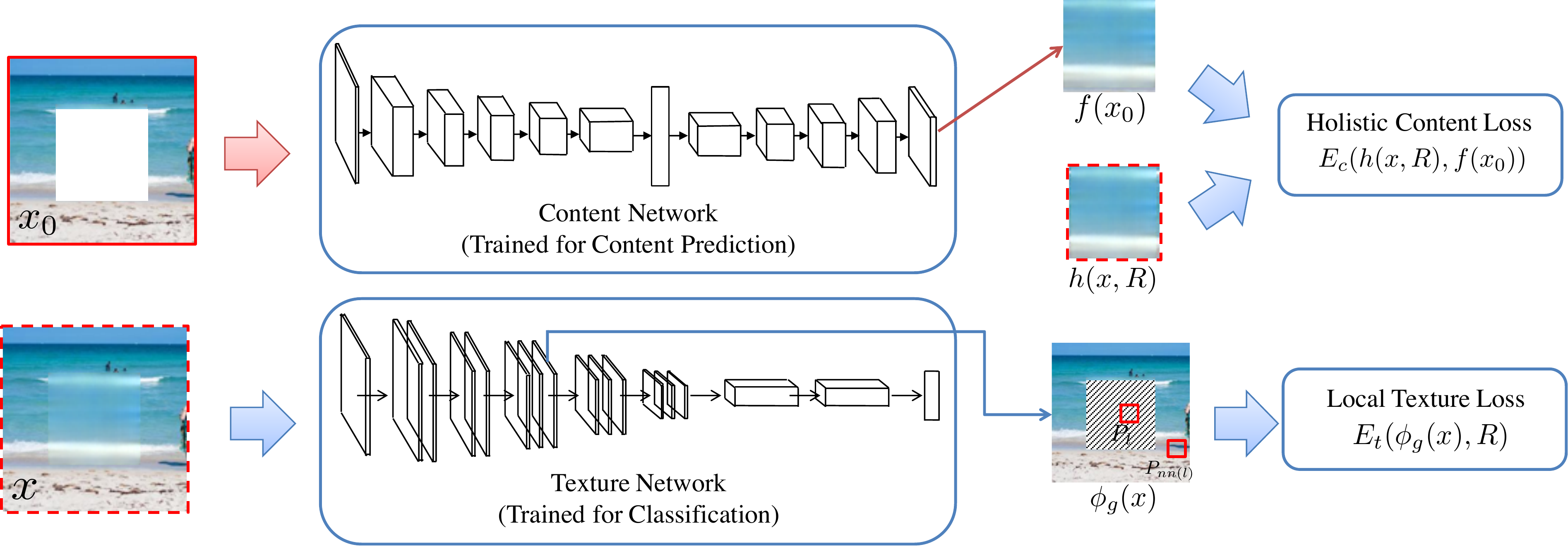}
	\caption{Framework Overview. Our method solves for an unknown image $x$ using two loss functions, the holistic content loss ($E_c$) and the local texture loss ($E_t$).
		At the smallest scale, the holistic content loss is conditioned on the output of the pre-trained content network given the input $x_0$ ($f(x_0)$).
		The local texture loss is derived by feeding $x$ into a pre-trained network (the texture network) and comparing the local neural patches between $R$ (the hole) and the boundary. \label{fig:framework}} 
\end{figure*}

\noindent\textbf{Structure Prediction using Deep Networks}
Over the recent years, convolutional neural networks have significantly advanced the image classification performance, as presented in~\cite{alexnet,googlenet,vggnet,residue152net}.
Meanwhile, researchers use deep neural networks for structure prediction~\cite{fcn,deeplab,deconvseg,eyescream,lstm_imggen,gan,draw,rnn_adnet,invertnet15dec,pixelrnn}, semantic segmentation~\cite{fcn,deeplab,deconvseg}, and image generation~\cite{gan,draw,eyescream,pixelrnn}. We are motivated by the generative power of deep neural network and use it as the backbone of our hole-filling approach. Unlike the image generation tasks discussed in~\cite{generatechair,gan,draw,eyescream}, where the input is a random noise vector and the output is an image, our goal is to predict the content in the hole, conditioned on the \emph{known image regions}.
Recently,~\cite{philip} proposed an encoder-decoder network for image inpainting, using the combination of the $\ell_2$ loss and the adversarial loss (Context Encoder).
In our work, we adapt Context Encoder as the global content prediction network and use the output to initialize our multi-scale neural patch synthesis algorithm at the smallest scale.

\noindent\textbf{Style Transfer}
In order to create realistic image textures, our work is motivated by the recent success of neural style transfer~\cite{gatys,gatysnips,chuanli,compositiontransfer,texturenetwork,percepturalloss}. These approaches are largely used to generate images combining the ``style'' of one image and the ``content'' of another image. Our technique is motivated by the astounding performance of neural style transfer. In particular, we show neural features are also extremely powerful to create fine textures and high-frequency details of natural images.

\section{The Approach}\label{sec:approach}

\subsection{Framework Overview}
We seek an inpainted image $\tilde{x}$ that optimizes over the loss function, which is formulated as a combination of three terms: the holistic {\em content} term, the local {\em texture} term, and the {\em tv-loss} term. The content term is a global structure constraint that captures the semantics and the global structure of the image, and the texture term models the local texture statistics of the input image. We first train the content network and use it to initialize the content term. The texture term is computed using the VGG-19 network~\cite{vgg}(Figure~\ref{fig:framework}) pre-trained on ImageNet.

To model the content constraint, we first train the holistic \textbf{content network} $f$. The input is an image with the central squared region removed and filled with the mean color, and the ground truth image $x_t$ is the original content in the center. We trained on two datasets, as discussed in Section~\ref{sec:exp}. Once the content network is trained, we can use the output of the network $f(x_0)$ as the initial content constraint for joint optimization.

The goal of the texture term is to ensure that the fine details in the missing hole are similar to the details outside of the hole. We define such similarity with \emph{neural patches}, which have been successfully used in the past to capture image styles. In order to optimize the texture term, we feed the image $x$ into the pre-trained VGG network (we refer to this network as local \textbf{texture network} in this paper) and enforce that the response of the small (typically $3 \times 3$) neural patches inside the hole region are similar to neural patches \emph{outside} the hole at pre-determined feature layers of the network. In practice we use the combination of \emph{relu3\_1} and \emph{relu4\_1} layers to compute the neural features. We iteratively update $x$ by minimizing the joint content and texture loss using limited-memory BFGS.

The proposed framework naturally applies to the high-resolution image inpainting problem using multiscale scheme. Given a high-resolution image with a large hole, we first downsize the image and obtain a reference content using the prediction of the content network. Given the reference content we optimize w.r.t. the content and texture constraints at the low resolution. The optimization result is then upsampled and used as the initialization for joint optimization at the fine scales. In practice, we set the number of scales to be 3 for images of size $512\times 512$. 

We describe the details of the three loss terms in the following.

\subsection{The Joint Loss Function}
\begin{figure*}[t]
	\centering
	\includegraphics[width=0.85\linewidth]{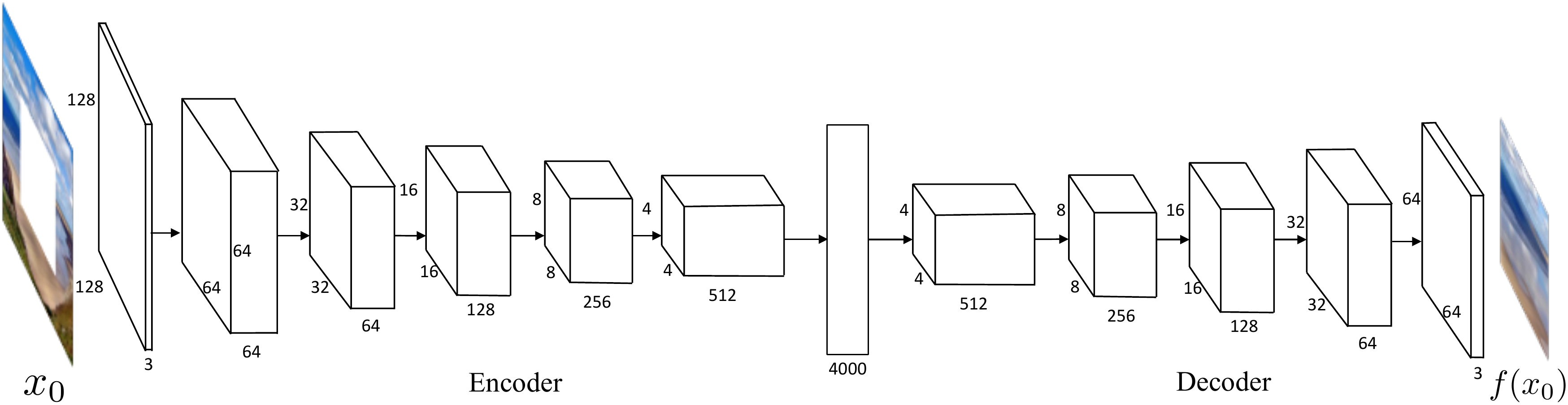}
	\caption{The network architecture for structured content prediction. Unlike the \(\ell_2\) loss architecture presented in~\cite{philip}, we replaced all ReLU/ReLU leaky layers with the ELU layer~\cite{elu} and adopted fully-connected layers instead of channel-wise fully-connected layers. The ELU unit makes the regression network training more stable than the ReLU leaky layers as it can handle large negative responses during the training process.}\label{fig:arch_content}

\end{figure*}

Given the input image $x_0$ we would like to find the unknown output image $x$. We use $R$ to denote a hole region in $x$, and $R^{\phi}$ to denote the corresponding region in a feature map $\phi(x)$ of the VGG-19 network.
$h(\cdot)$ defines the operation of extracting a sub-image or sub-feature-map in a rectangular region, i.e. $h(x,R)$ returns the color content of $x$ within $R$, and $h(\phi(x),R^{\phi})$ returns the content of $\phi(x)$ within $R^{\phi}$, respectively. We denote the content network as $f$ and the texture network as $t$.

At each scale $i=1,2...,N$ ($N$ is the number of scales), the optimal reconstruction (hole filling) result
$\tilde{x}$ is obtained by solving the following minimization problem:
\begin{eqnarray}
\tilde{x}_{i+1} = && \argmin_{x} E_c(h(x,R), h(x_i,R)) \nonumber \\ 
&& +\alpha E_t(\phi_t(x),R^{\phi})  + \beta \Upsilon(x)
\label{eq:opt}
\end{eqnarray}
where $h(x_1,R)=f(x_0)$, $\phi_t(\cdot)$ represents a feature map (or a combination of feature maps) at an intermediate layer in the texture network $t$, and $\alpha$ is a weight reflecting the importance between the two terms. Empirically setting $\alpha$ and $\beta$ to be $5\mathrm{e}^{-6}$ balances the magnitude of each loss and gives best results in our experiment.

The first term $E_c$ in Equation \ref{eq:opt} which models \textbf{the holistic content constraint} is defined to penalize the $\ell_2$ difference between the optimization result and the previous content prediction (from the content network or the result of optimization at the coarser scale):
\begin{equation}
E_c(h(x,R), h(x_i,R))  = \parallel h(x,R) - h(x_i,R) \parallel_2^2
\end{equation}

The second term $E_t$ in Equation \ref{eq:opt} models \textbf{the local texture constraint}, which penalizes the discrepancy of the texture appearance inside and outside the hole.
We first choose a certain feature layer (or a combination of feature layers) in the network $t$, and extract its feature map $\phi_t$.
For each local query patch $P$ of size $s \times s \times c$ in the hole $R^{\phi}$, we find its most similar patch outside the hole, and compute the loss by averaging the distances of the query patch and its nearest neighbor.
\begin{multline}
E_t(\phi_t(x),R)  = \\ 
\frac{1}{|R^{\phi}|}\sum_{i\in R^{\phi}} \parallel h(\phi_t(x),P_i) - h(\phi_t(x),P_{nn(i)})  \parallel_2^2
\end{multline}
where $|R^{\phi}|$ is the number of patches sampled in the region $R^{\phi}$, $P_{i}$ is the local neural patch centered at location $i$, and $nn(i)$ is the computed as
\begin{equation}
nn(i) = \argmin_{j\in \mathcal{N}(i) \wedge  j \notin R^{\phi} } \parallel h(\phi_t(x),P_i) - h(\phi_t(x),P_j) \parallel_2^2
\end{equation}
where $\mathcal{N}(i)$ is the set of neighboring locations of $i$ excluding the overlap with $R^{\phi}$.
The nearest neighbor can be fast computed as a convolutional layer, as shown in~\cite{chuanli}.

We also add the TV loss term to encourage smoothness:
\begin{equation}
\Upsilon(x)  = \sum_{i,j}((x_{i,j+1} - x_{i,j})^2 + (x_{i+1,j} - x_{i,j})^2)
\end{equation}


\subsection{The Content Network}~\label{sec:content}
A straightforward way to learn the initial content prediction network is to train a regression network $f$ to use the response $f(x)$ of an input image $x$ (with the unknown region) to approximate the ground truth $x_g$ at the region $R$.
Recent studies have used various loss functions for image restoration tasks, for instance, \(\ell_2\) loss, SSIM loss~\cite{zhao:nnip,alexey:ssim,karl:ssim}, \(\ell_1\) loss~\cite{zhao:nnip}, perceptual loss~\cite{percepturalloss}, and adversarial loss~\cite{philip}.
We experimented with \(\ell_2\) loss and adversarial loss. For each training image, the \(\ell_2\) loss is defined as:
\begin{equation}
L_{l2}(x, x_g, R) = \parallel f(x) - h(x_g, R) \parallel_2^2
\end{equation}

The adversarial loss is defined as: 
\begin{eqnarray}
L_{adv}(x, x_g, R) = \max_{D} E_{x \in \mathcal{X}}[\log(D(h(x_g, R)))  \nonumber \\ 
 +\log(1-D(f(x)))]
\end{eqnarray}
where D is the adversarial discriminator.

We use the joint \(\ell_2\) loss and the adversarial loss the same way as the Context Encoder~\cite{philip}:

\begin{equation}
L = \lambda L_{l2}(x, x_g, R) + (1 - \lambda) L_{adv}(x, x_g, R)
\end{equation}
where $\lambda$ is $0.999$ in our implementation.

\subsection{The Texture Network}
We use the VGG-19\cite{vgg} network pre-trained for ImageNet classification as the texture network, and use the \emph{relu3\_1} layer and the \emph{relu4\_1} layer to calculate the texture term. We found using a combination of \emph{relu3\_1} and \emph{relu4\_1} leads to more accurate results than using a single layer. As an alternative, we tried to use the content network discussed in the previous section as the texture network, but found the results are of lower quality than using the pre-trained VGG-19. This can be explained by the fact that the VGG-19 network was trained for semantic classification, so features of its intermediate layers manifest strong invariance w.r.t. texture distortions. This helps infer more accurate reconstruction of the hole content.



\section{Experiments}\label{sec:exp}
This section evaluates our proposed approach visually and quantitatively. We first introduce the datasets and then compare our approach with other methods, demonstrating its effectiveness in high-resolution image inpainting. At the end of this section we show a real world application where we remove distractors from photos.

\noindent\textbf{Datasets} 
We evaluate the proposed approach on two different datasets: Paris StreetView~\cite{doersch2012makes} and ImageNet~\cite{russakovsky2015imagenet}. Labels or other information associated with these images are not used.  The Paris StreetView contains 14,900 training images and 100 test images. ImageNet has 1,260,000 training images, and 200 test images that are randomly picked from the validation set. We also picked 20 images with distractors to test out our algorithm for distractor removal.

\noindent\textbf{Experimental Settings} We first compare our method with several baseline methods in the low-resolution setting ($128\times 128)$. First, we compared with results of Context Encoder trained with \(\ell_2\) loss. Second, we compare our method with the best results that Context Encoder have achieved using adversarial loss, which is the state-of-the-art in the area of image inpainting using deep learning. Finally, we compare with the results of Content-Aware Fill using PatchMatch algorithm from Adobe Photoshop. Our comparisons demonstrate the effectiveness of the proposed joint optimization framework. 

While comparisons with baselines show the effectiveness of the overall joint optimization algorithm and the role of the texture network in joint optimization, we further analyze the separate role of the content network and the texture network by changing their weights in the joint optimization.   

Finally, we show our results on high-resolution image inpainting and compare with Content-Aware Fill and Context Encoder (\(\ell_2\) and adversarial loss). Note that for Context Encoder the high-resolution results are acquired by directly upsampling from the low-resolution outputs. Our approach shows significant improvement in terms of the visual quality.

\begin{figure}[h!]
  \centering
  \includegraphics[width=0.95\linewidth]{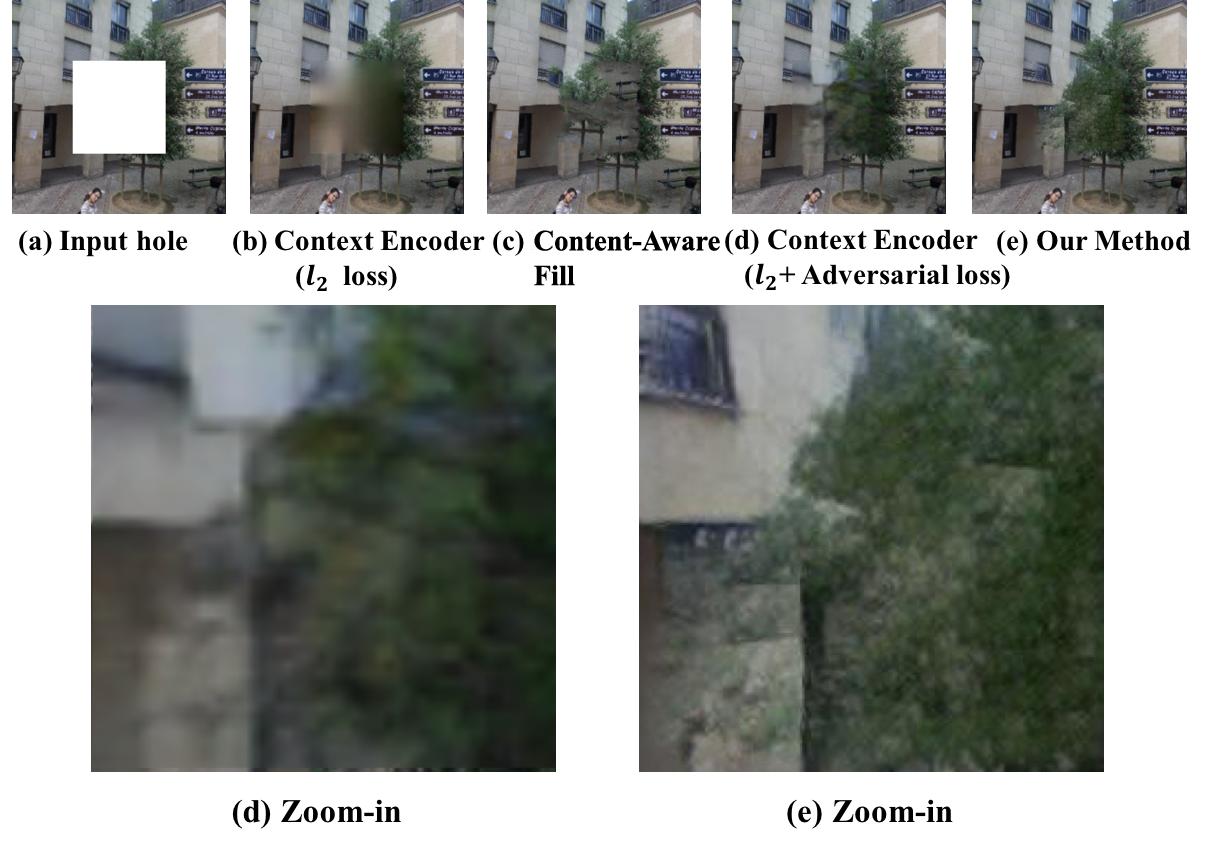}
  \caption{Comparison with Context Encoder (\(\ell_2\) loss), Context Encoder (\(\ell_2\) loss + adversarial loss) and Content-Aware Fill. We can see that our approach fixes the wrong textures generated by Content-Aware Fill, and is also more clear than the output of Context Encoder.} 
  \label{fig:low_res}
 
\end{figure}

\noindent\textbf{Quantitative Comparisons}
We first compare our method quantitatively with the baseline methods on low-resolution images ($128\times 128$) on the Paris StreetView dataset. Results in Table~\ref{table:paris} show that our method achieves highest numerical performance. We attribute this to the nature of our method -- it can infer the correct structure of the image where Content-Aware Fill fails, and can also synthesize better image details comparing with the results of Context Encoder (Fig.~\ref{fig:low_res}). In addition, we argue that the quantitative evaluation may not be most effective measure of the inpainting task given that the goal is to generate realistic-looking content, rather than exact same content that was in the original image. 

\begin{table}[h!]
\begin{center}
\resizebox{.45\textwidth}{!}{%
  \begin{tabular}{ l  c  c c}
    \hline
    Method & Mean L1 Loss & Mean L2 Loss & PSNR \\ \hline
    \emph{Context Encoder \(\ell_2\) loss} & 10.47\% &  2.41\% & 17.34 dB\\ \hline
    \emph{Content-Aware Fill} & 12.59\% & 3.14\% & 16.82 dB\\ \hline
    \emph{Context Encoder (\(\ell_2\) + adversarial loss)} & 10.33\% & 2.35\% & 17.59 dB\\ \hline
    \emph{Our Method} & \textbf{10.01}\% & \textbf{2.21}\% & \textbf{18.00 dB} \\ \hline
    \hline
  \end{tabular}}
  \end{center}
  \caption{Numerical comparison on Paris StreetView dataset. Higher PSNR value is better. Note \% in the Table is to facilitate reading.}
  \label{table:paris}
 
\end{table}

\begin{figure*}[t]
  \center
\setlength\tabcolsep{1.5pt}
\begin{tabular}{ccccc}
  \includegraphics[width=.2\textwidth]{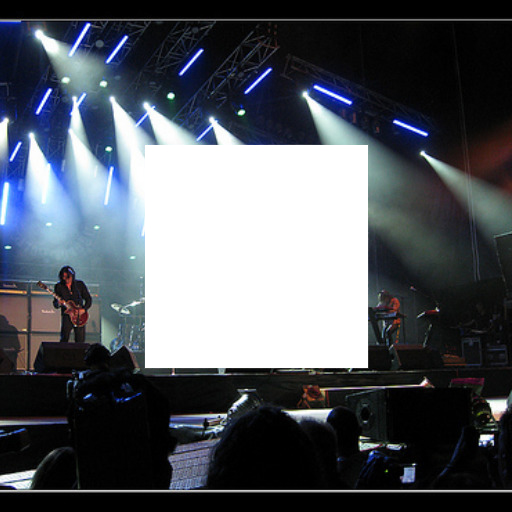}&
  \includegraphics[width=.2\textwidth]{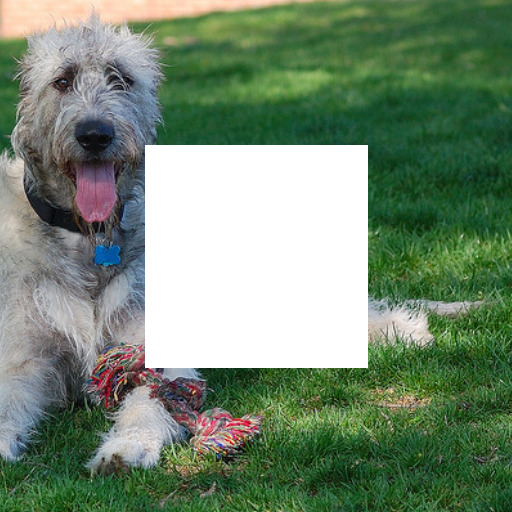}&
  \includegraphics[width=.2\textwidth]{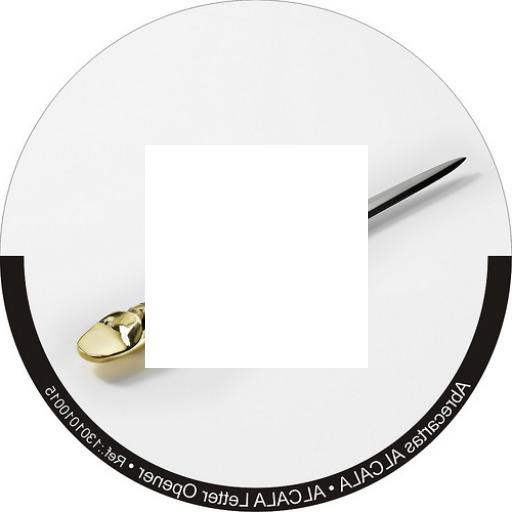}&
  \includegraphics[width=.2\textwidth]{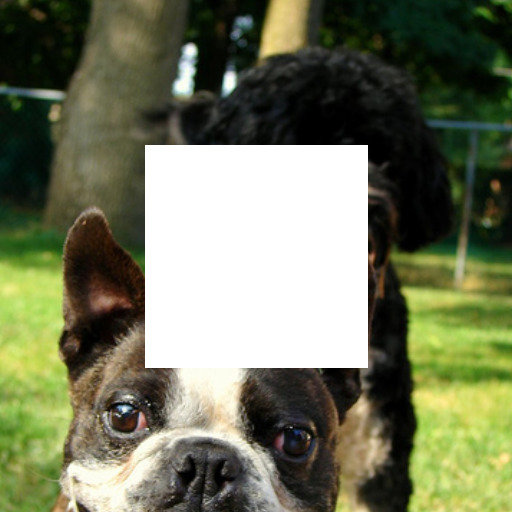}&
  \includegraphics[width=.2\textwidth]{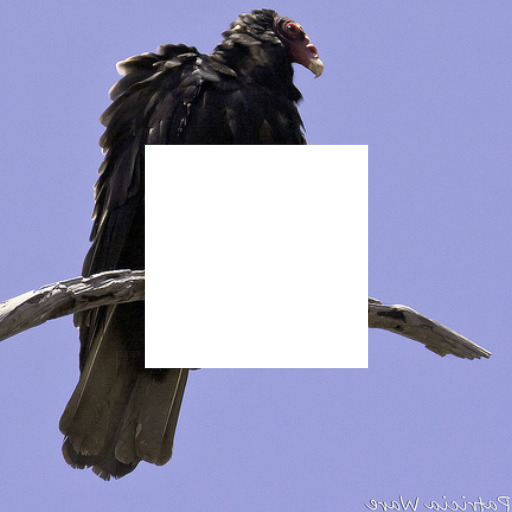}\\

  \includegraphics[width=.2\textwidth]{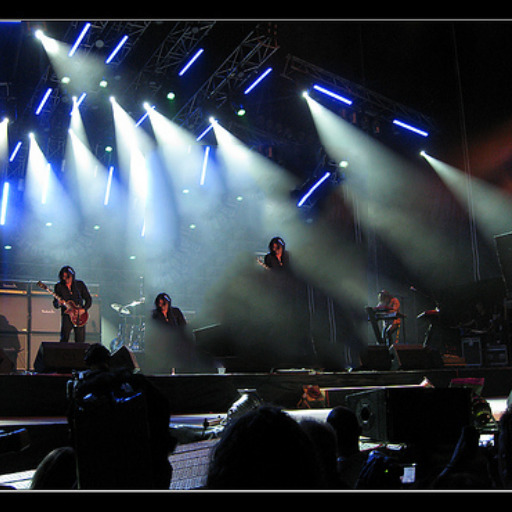}&
  \includegraphics[width=.2\textwidth]{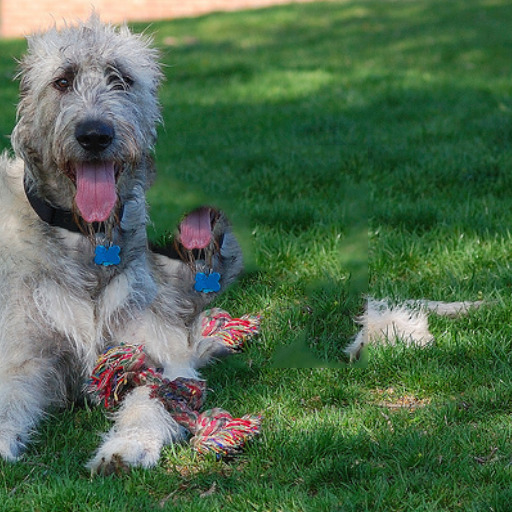}&
  \includegraphics[width=.2\textwidth]{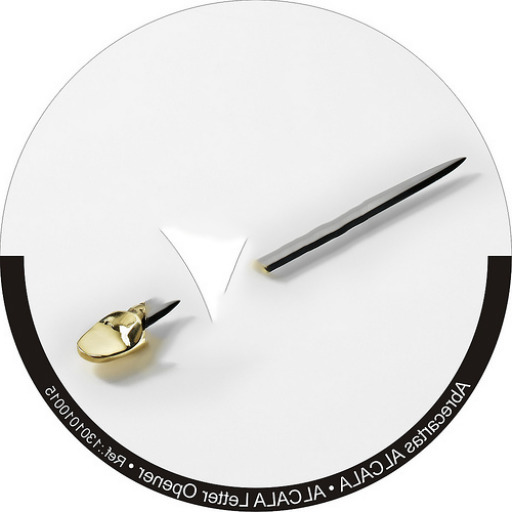}&
  \includegraphics[width=.2\textwidth]{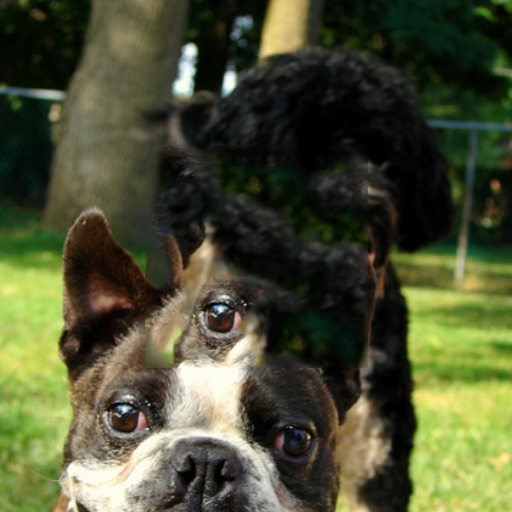}&
  \includegraphics[width=.2\textwidth]{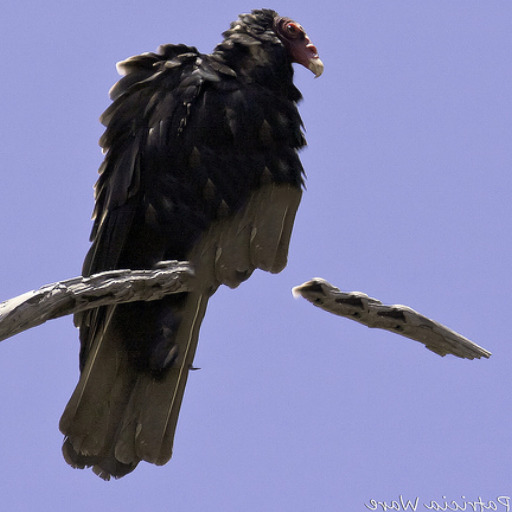}\\

  \includegraphics[width=.2\textwidth]{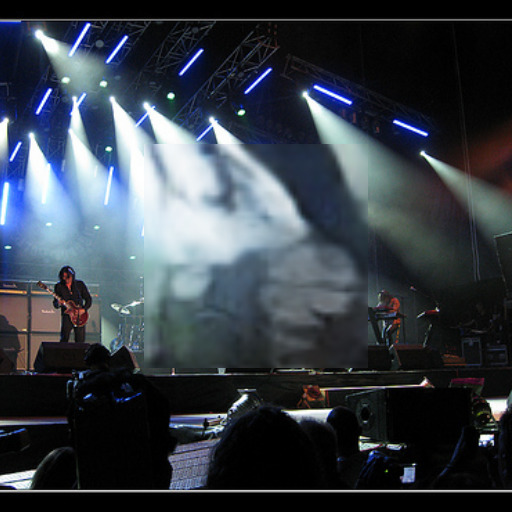}&
  \includegraphics[width=.2\textwidth]{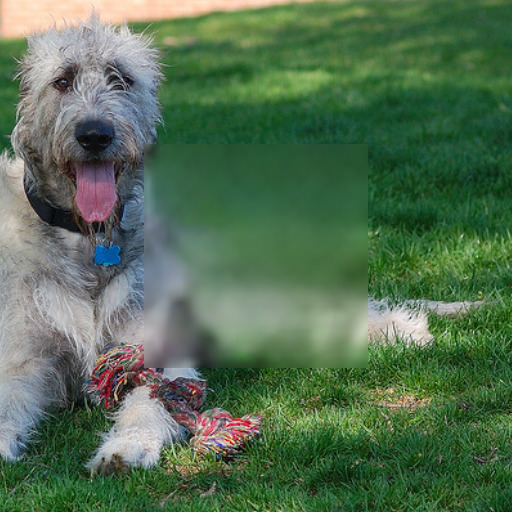}&
  \includegraphics[width=.2\textwidth]{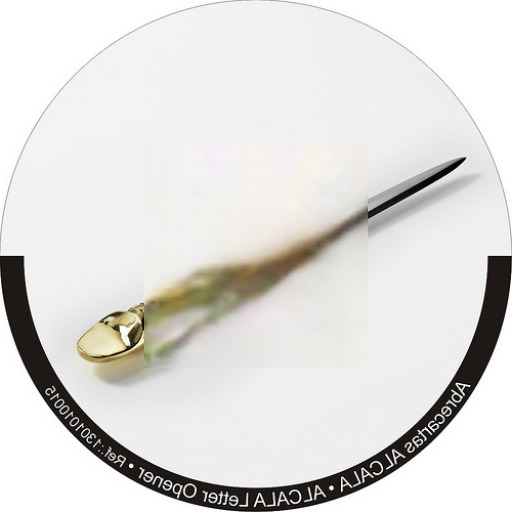}&
  \includegraphics[width=.2\textwidth]{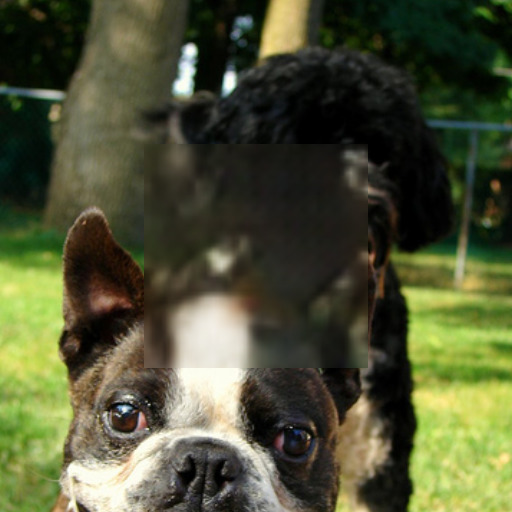}&
  \includegraphics[width=.2\textwidth]{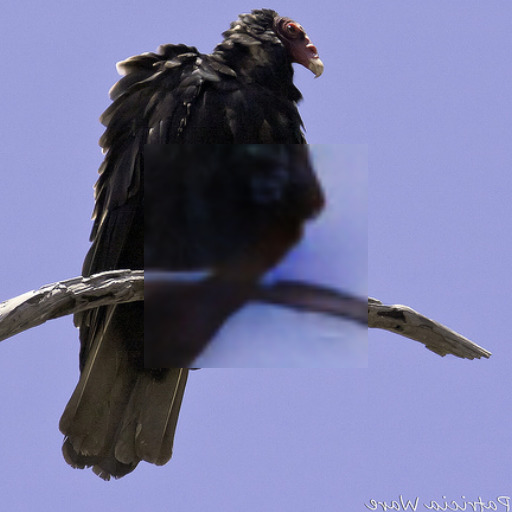}\\

  \includegraphics[width=.2\textwidth]{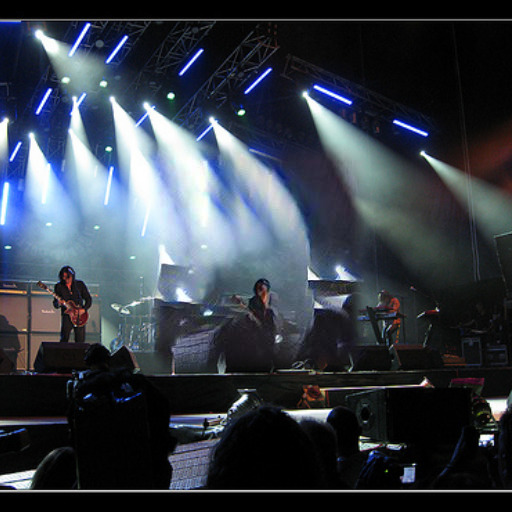}&
  \includegraphics[width=.2\textwidth]{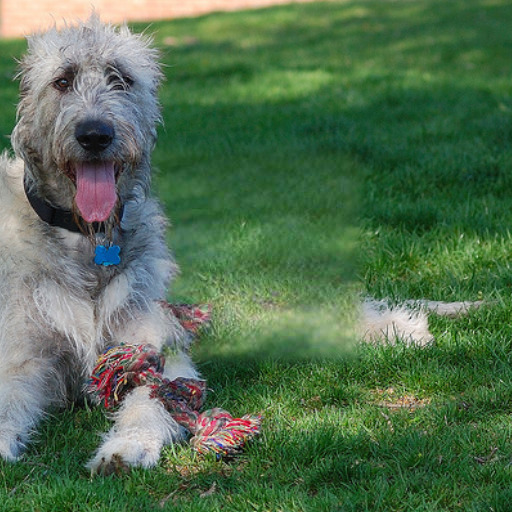}&
  \includegraphics[width=.2\textwidth]{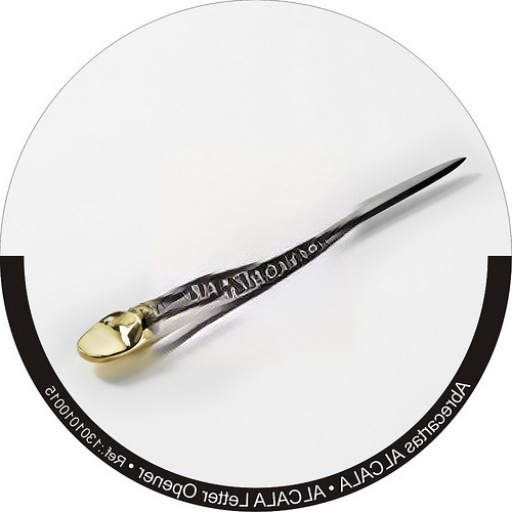}&
  \includegraphics[width=.2\textwidth]{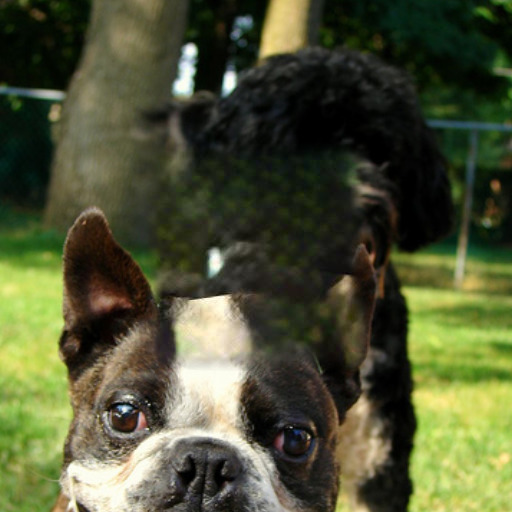}&
  \includegraphics[width=.2\textwidth]{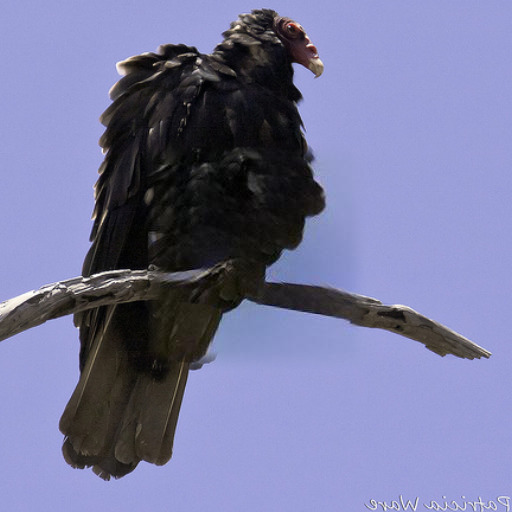} \\

\end{tabular}
\caption{Visual comparisons of ImageNet result. From top to bottom: input image, Content-Aware Fill, Context Encoder (\(\ell_2\) and adversarial loss), our result. All images are scaled from $512\times 512$ to fit the page size.}
\label{fig:imagenet}
\end{figure*} 

\begin{figure}[h!]
\setlength\tabcolsep{1.5pt}
\centering
\small
\begin{tabular}{cccc}
\includegraphics[width=.24\linewidth]{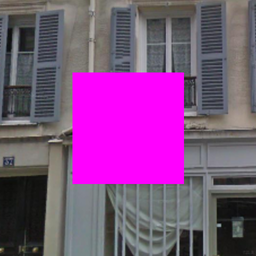} &
\includegraphics[width=.24\linewidth]{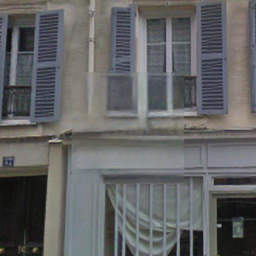} &
\includegraphics[width=.24\linewidth]{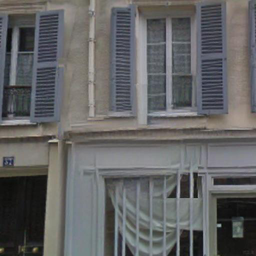} &
\includegraphics[width=.24\linewidth]{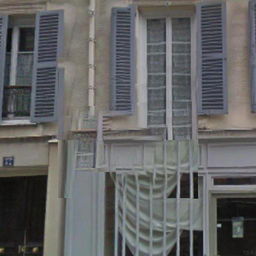} \\
(a) Input image & (b) $\alpha=1e-6$ & (c) $\alpha=1e-5$ & (d) $\alpha=4e-5$\\ 
\end{tabular}
\caption{The effect of different texture weight $\alpha$. }\label{alpha}
\label{fig:alpha}

\end{figure}

\begin{figure}[h!]
\setlength\tabcolsep{1.5pt}
\centering
\begin{tabular}{cccc}

\includegraphics[width=.16\textwidth]{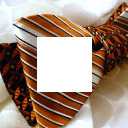}&
\includegraphics[width=.16\textwidth]{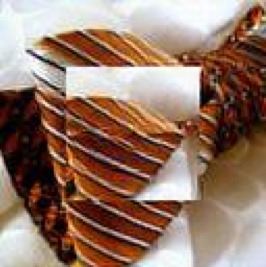} &
\includegraphics[width=.16\textwidth]{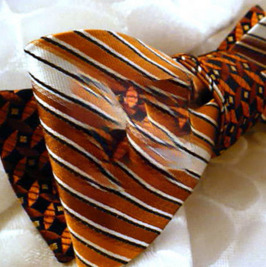} \\
(a) & (b) & (c)  \\ 
\end{tabular}
\caption{Evaluation of different components. (a) input image. (b) result without using content constraint. (c) our result. }\label{fig:ablation}

\end{figure}

\noindent\textbf{The effects of content and texture networks} 
One ablation study we did was to drop the content constraint term and only use the texture term in the joint optimization. As shown in Fig.~\ref{fig:ablation}, without using the content term to guide the optimization, the structure of the inpainting results is completely incorrect. We also adjusted the relative weight between the content term and the texture term. Our finding is that by using more content constraint, the result is more consistent with the initial prediction of the content network but may lack high frequency details. Similarly, using more texture term gives sharp result but does not guarantee the overall image structure is correct (Fig.~\ref{fig:alpha}).

\begin{figure}[h!]
  \setlength\tabcolsep{1.5pt}
  \centering
  \begin{tabular}{cccc}
    \includegraphics[width=.24\linewidth]{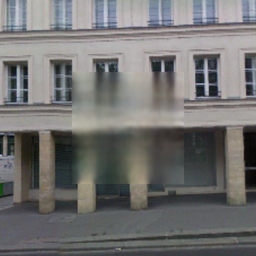} &
    \includegraphics[width=.24\linewidth]{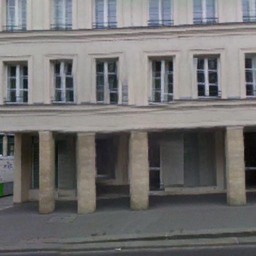} &
    \includegraphics[width=.24\linewidth]{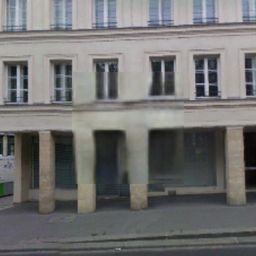} &
    \includegraphics[width=.24\linewidth]{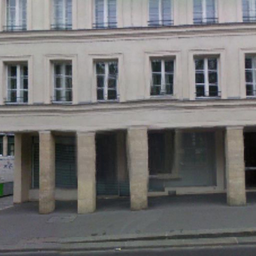} \\
    (a) & (b) & (c) & (d) \\ 

  \end{tabular}
  \caption{(a) Output of content network trained with $\ell_2$ loss (b) The final result using (a). (c) Output of content network trained with $\ell_2$ and adversarial loss. (d) The final result using (c).}\label{fig:adversarial}

\end{figure}

\noindent\textbf{The effect of the adversarial loss} 
We analyze the effect of using adversarial loss in training the content network. One may argue without using the adversarial loss, the content network is still able to predict the structure of the image and the joint optimization will calibrate the textures later. However we found that the quality of the initialization given by the content network is important to the final result. When the initial prediction is blurry (using $\ell_2$ loss only), the final result becomes more blurry as well comparing with using the content network trained with both $\ell_2$ and adversarial loss (Fig.~\ref{fig:adversarial}).

\begin{figure*}[t!]
	\center
\setlength\tabcolsep{1.5pt}
\begin{tabular}{ccccc}
  \includegraphics[width=.2\textwidth]{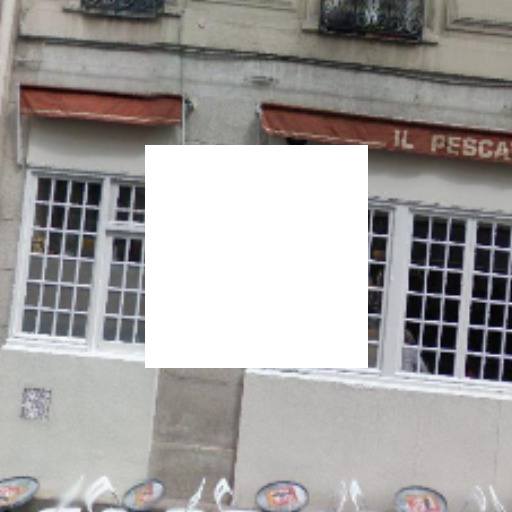}&
  \includegraphics[width=.2\textwidth]{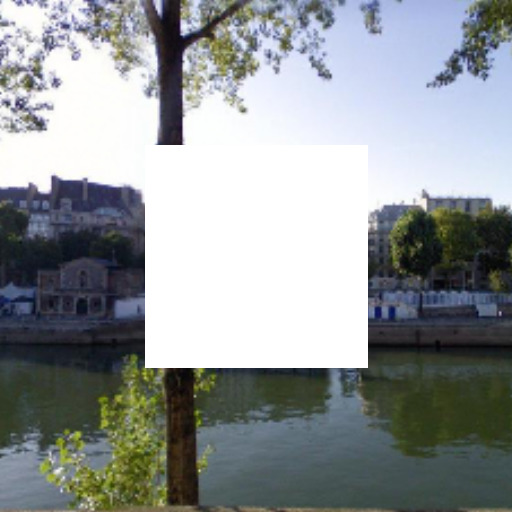}&
  \includegraphics[width=.2\textwidth]{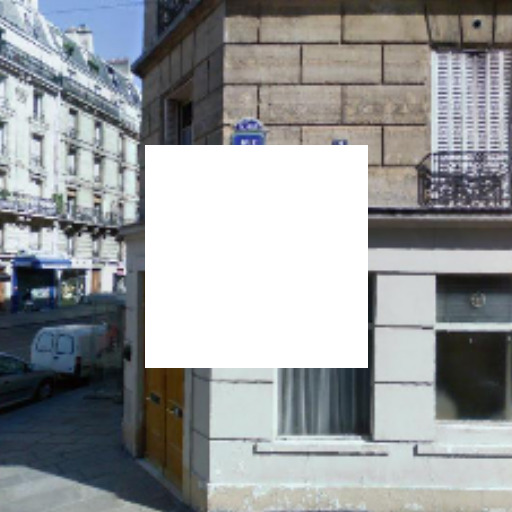}&
  \includegraphics[width=.2\textwidth]{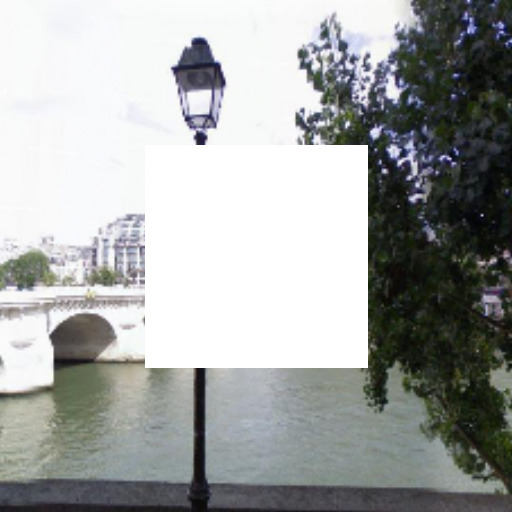}&
  \includegraphics[width=.2\textwidth]{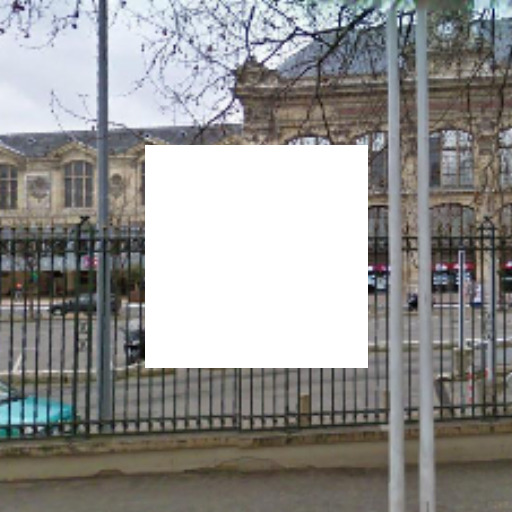}\\

  \includegraphics[width=.2\textwidth]{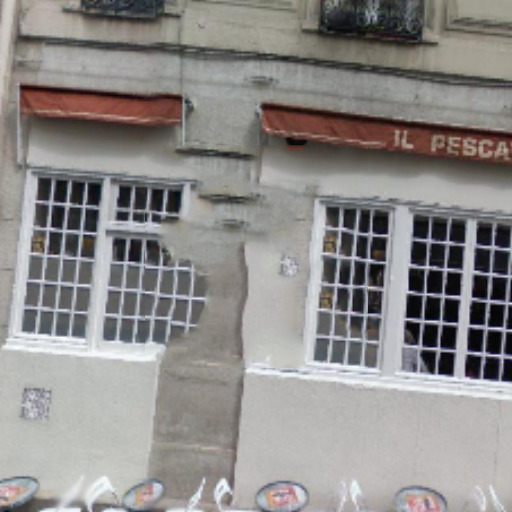}&
  \includegraphics[width=.2\textwidth]{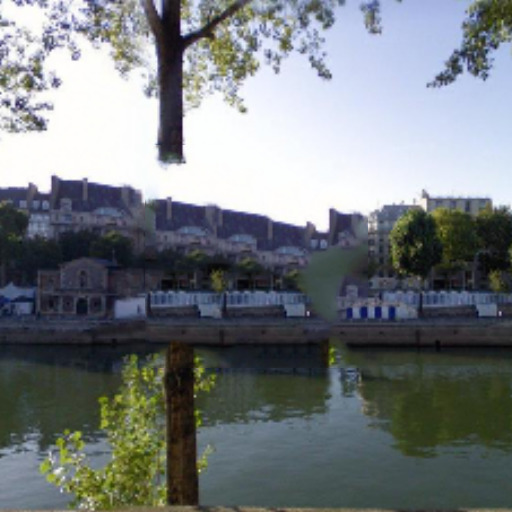}&
  \includegraphics[width=.2\textwidth]{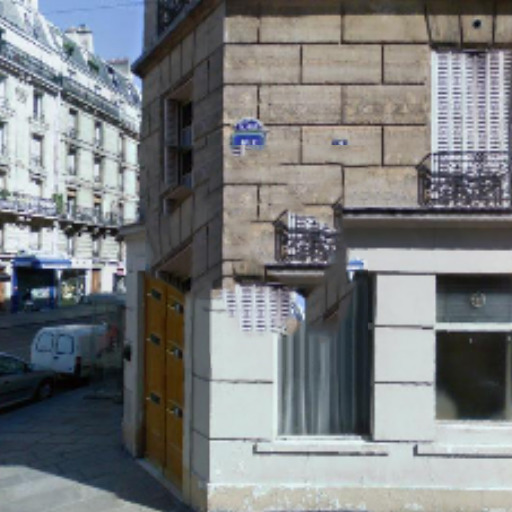}&
  \includegraphics[width=.2\textwidth]{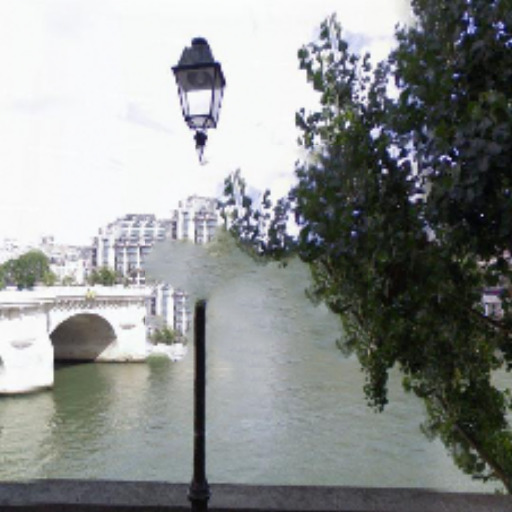}&
  \includegraphics[width=.2\textwidth]{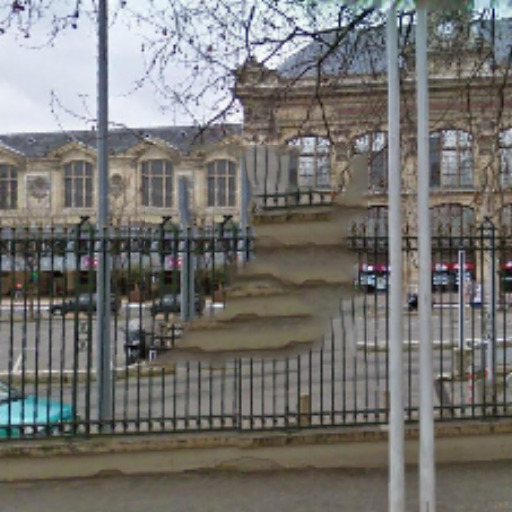}\\

  \includegraphics[width=.2\textwidth]{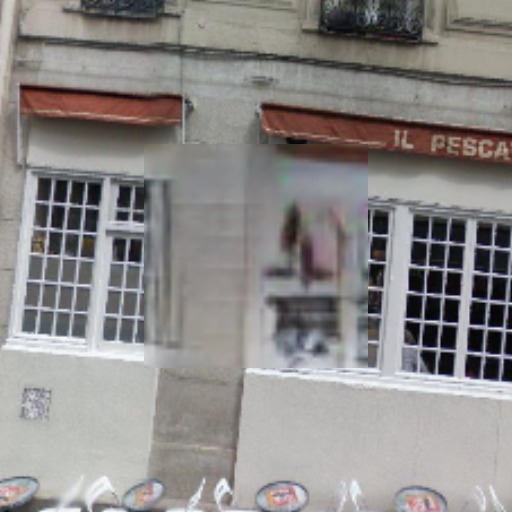}&
  \includegraphics[width=.2\textwidth]{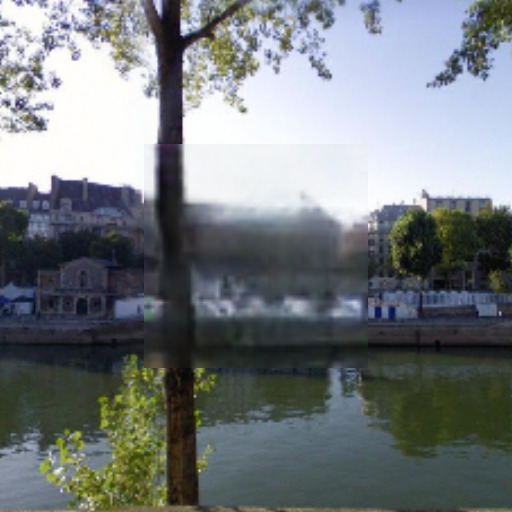}&
  \includegraphics[width=.2\textwidth]{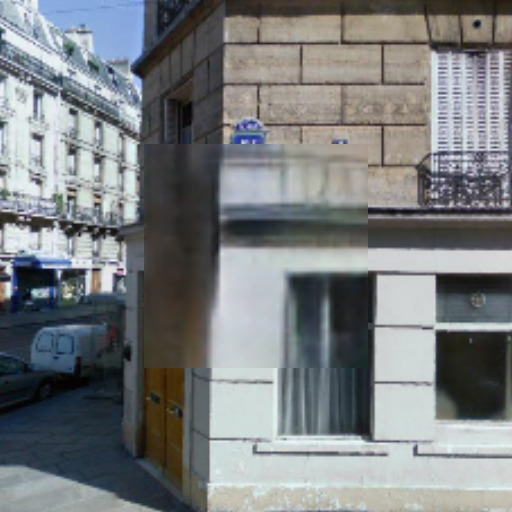}&
  \includegraphics[width=.2\textwidth]{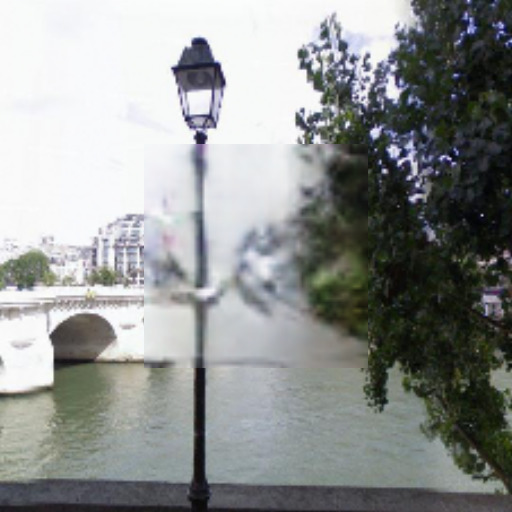}&
  \includegraphics[width=.2\textwidth]{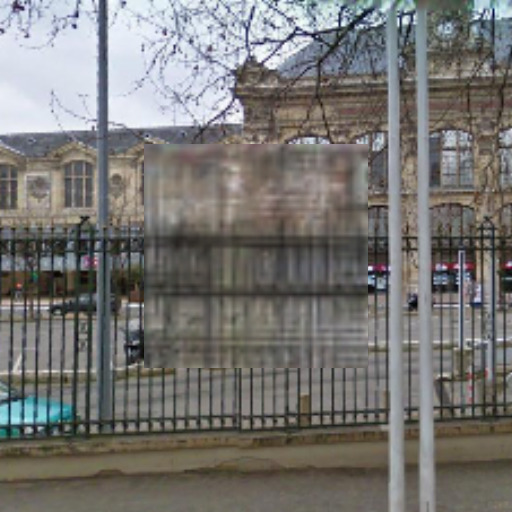}\\

  \includegraphics[width=.2\textwidth]{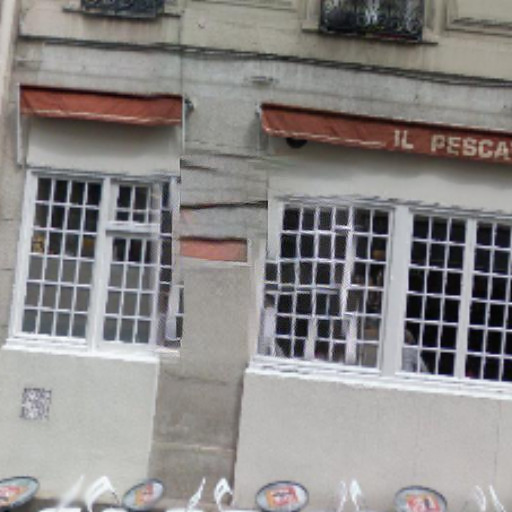}& 
  \includegraphics[width=.2\textwidth]{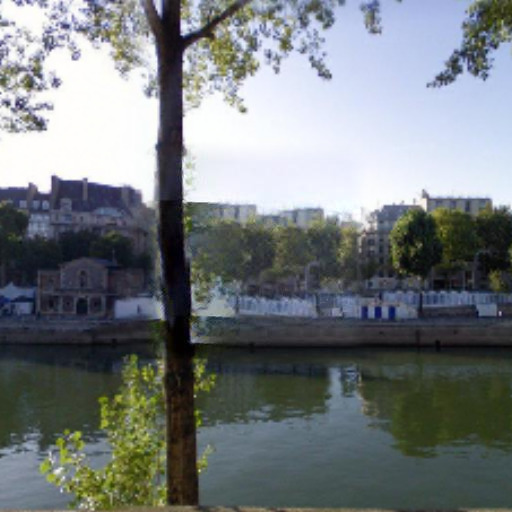}&
  \includegraphics[width=.2\textwidth]{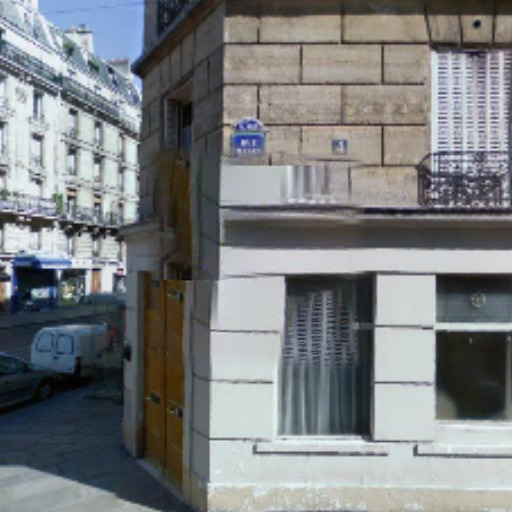}&
  \includegraphics[width=.2\textwidth]{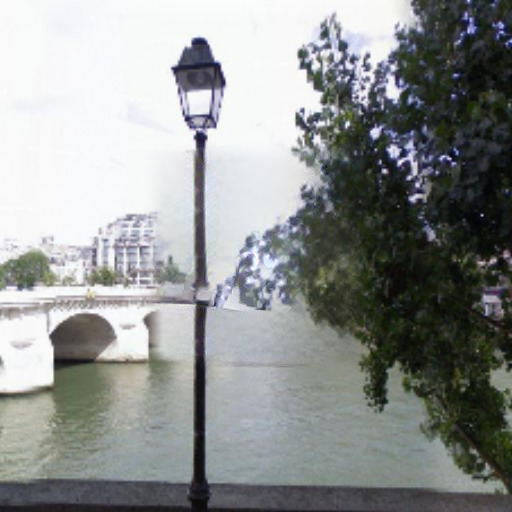}&
  \includegraphics[width=.2\textwidth]{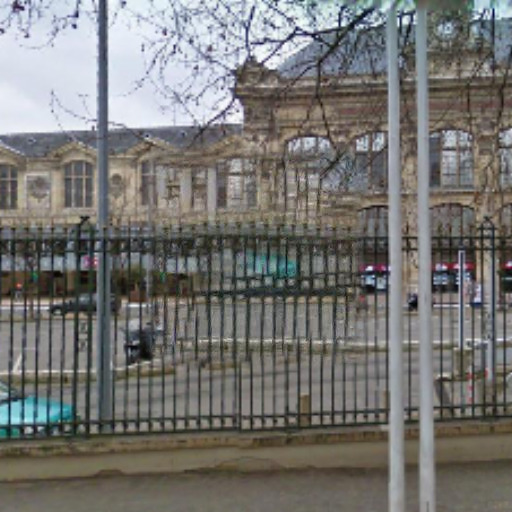} \\

\end{tabular}
\caption{Visual comparisons of Paris Streetview result. From top to bottom: input image, Content-Aware Fill, Context Encoder (\(\ell_2\) and adversarial loss) and our result. All images are scaled from $512\times 512$ to fit the page size.}

 \label{fig:paris}
\end{figure*}

\noindent\textbf{High-Resolution image inpainting}
We demonstrate our result of high-resolution image ($512\times512$) inpainting in Fig.~\ref{fig:imagenet} and Fig.~\ref{fig:paris} and compare with Content-Aware Fill and Context Encoder (\(\ell_2\) + adversarial loss). Since Context Encoder only works with 128x128 images and when the input is larger, we directly upsample the $128\times 128$ output to $512\times 512$ using bilinear interpolation. In most of the results, our multi-scale, iterative approach combines the advantage of the other approaches, producing results with coherent global structure as well as high-frequency details. As shown in figures, a significant advantage of our approach over Content-Aware Fill is that we are able to generate new textures as we do not propagate the existing patches directly. However, one disadvantage is that given our current implementation, our algorithm takes roughly $1$ min to fill in a $256 \times 256$ hole of a $512 \times 512$ image with a Titan X GPU, which is significantly slower than Content-Aware Fill.

\begin{figure}[!htb]
\setlength\tabcolsep{1.5pt}
\centering
\begin{tabular}{cccc}

\includegraphics[width=.12\textwidth]{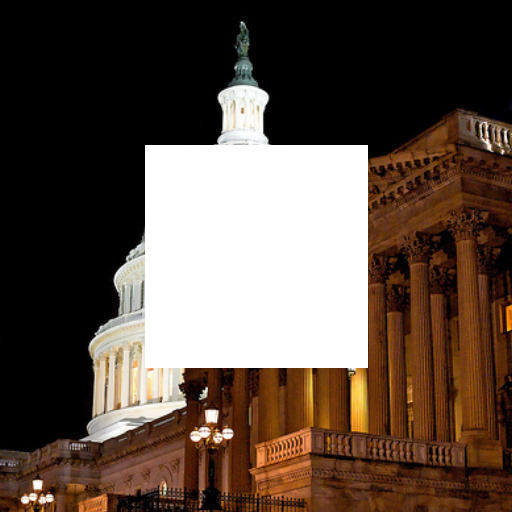}&
\includegraphics[width=.12\textwidth]{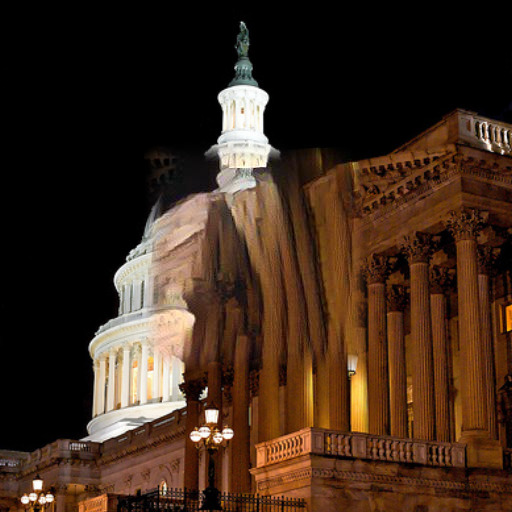}&
\includegraphics[width=.12\textwidth]{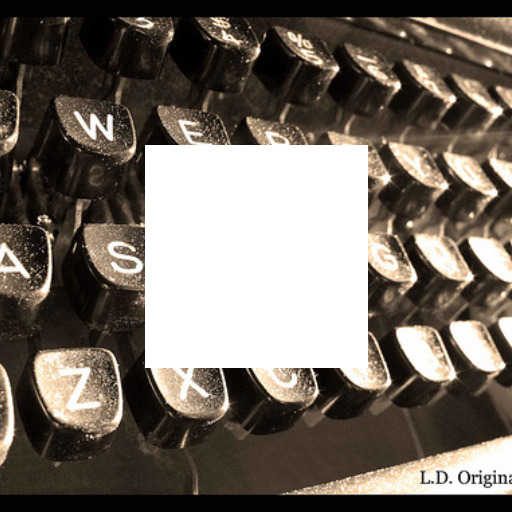}&
\includegraphics[width=.12\textwidth]{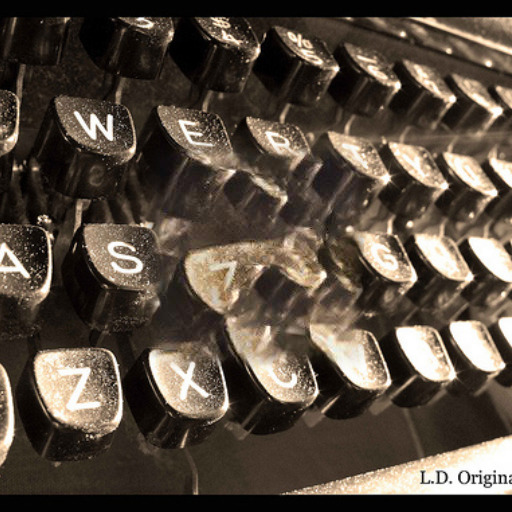}\\
  
\end{tabular}
\caption{Failure cases of our method.}

 \label{fig:failure}
\end{figure}

\begin{figure*}[!htb]
\setlength\tabcolsep{1.5pt}
\centering
\begin{tabular}{cccc}

\includegraphics[width=.23\textwidth]{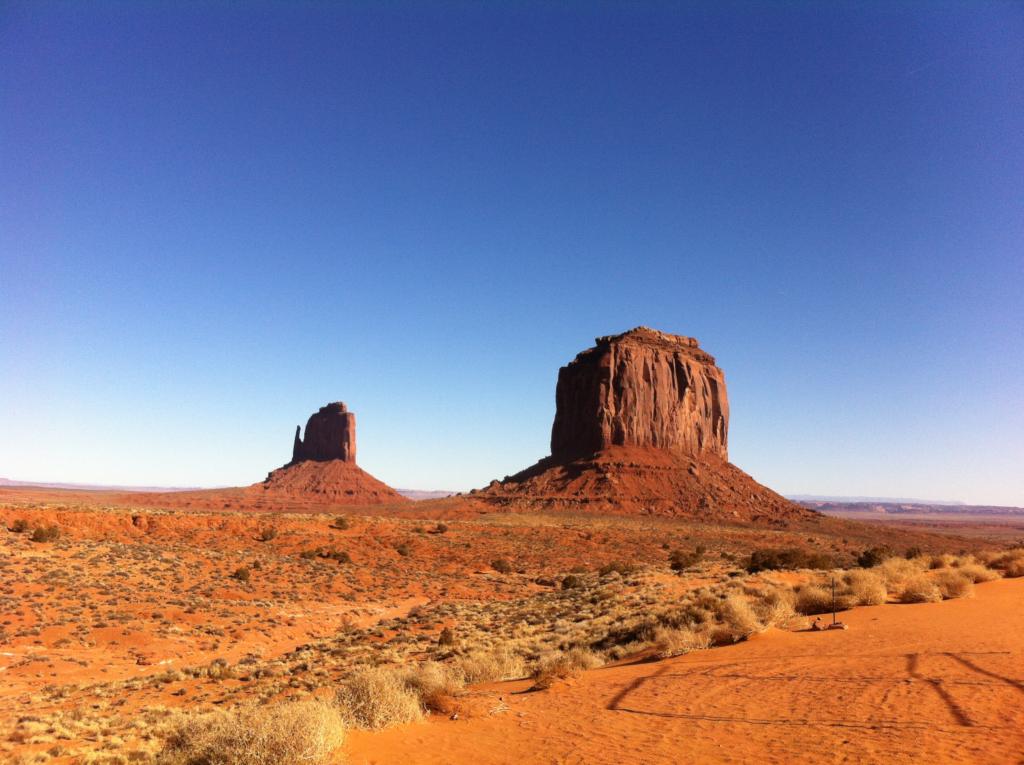}&
\includegraphics[width=.23\textwidth]{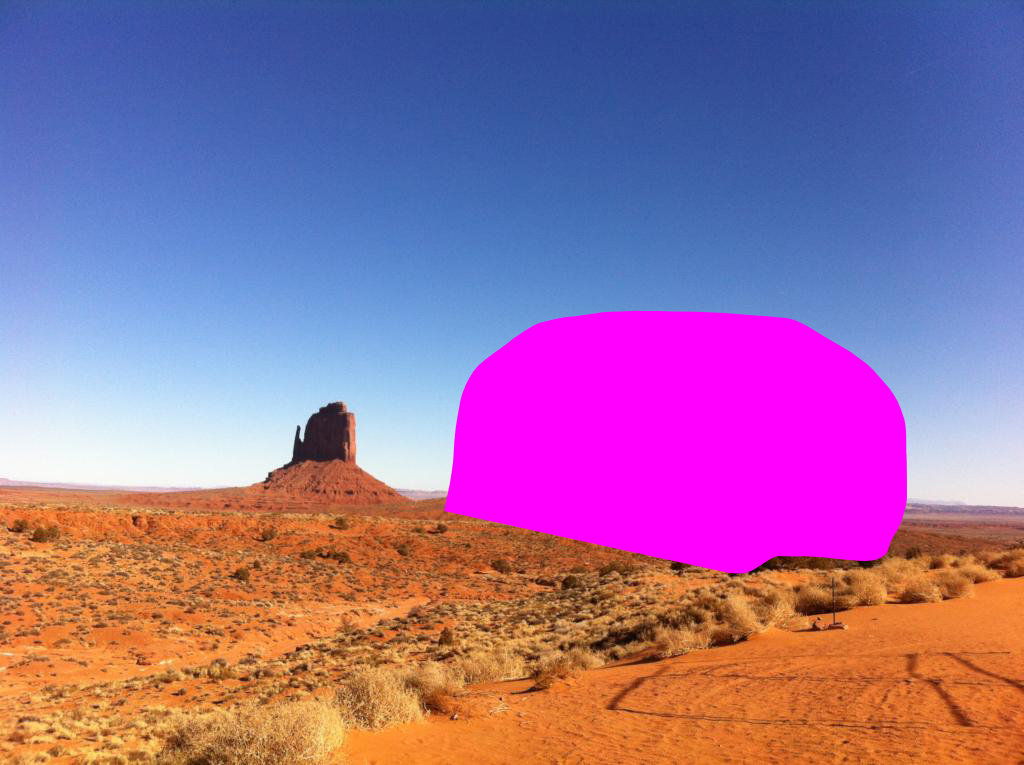}&
\includegraphics[width=.23\textwidth]{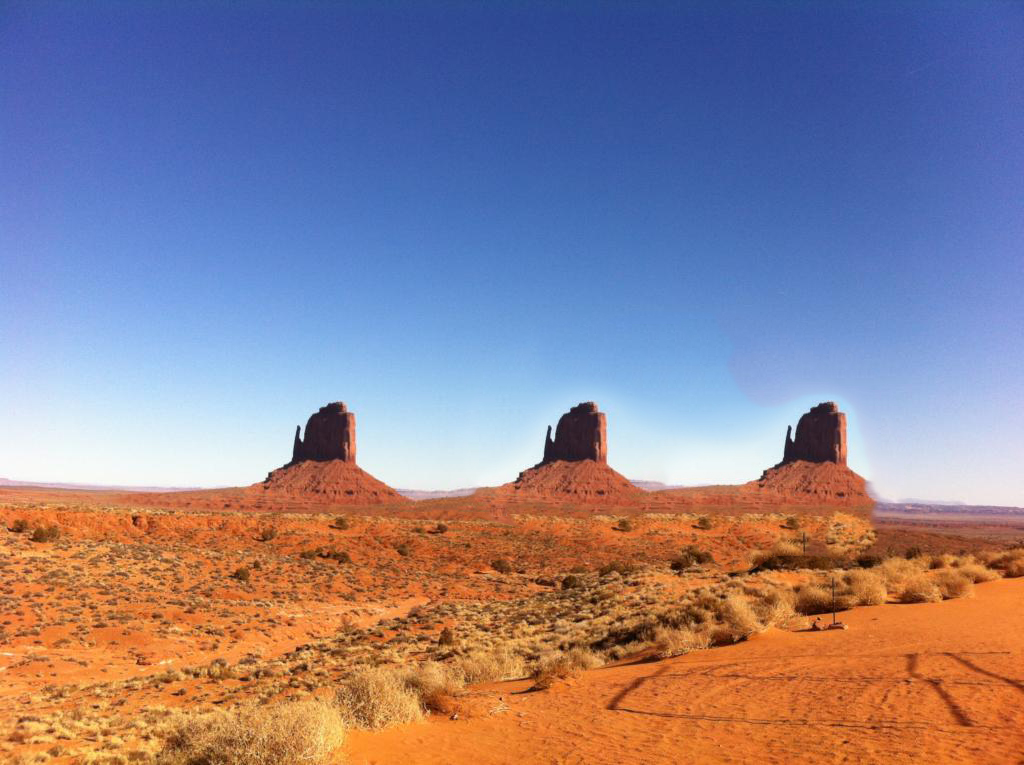}&
\includegraphics[width=.23\textwidth]{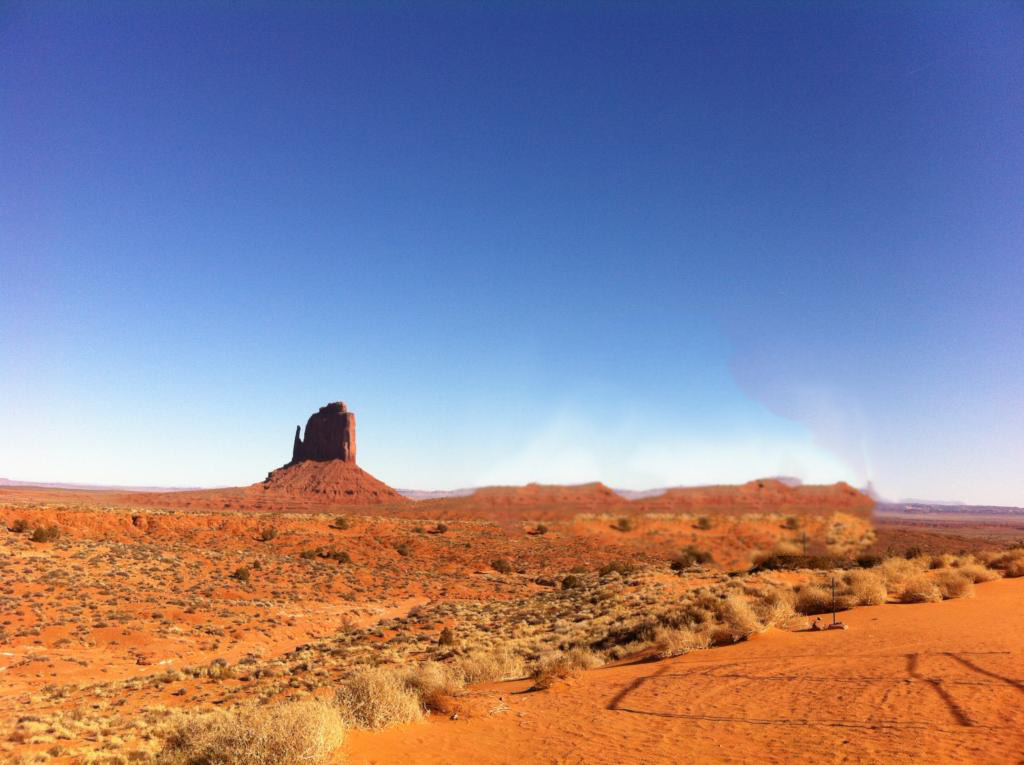} \\

\includegraphics[width=.23\textwidth, height=.15\textwidth]{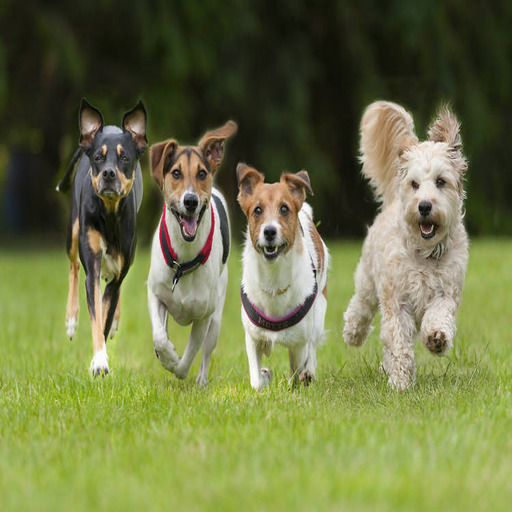}&
\includegraphics[width=.23\textwidth, height=.15\textwidth]{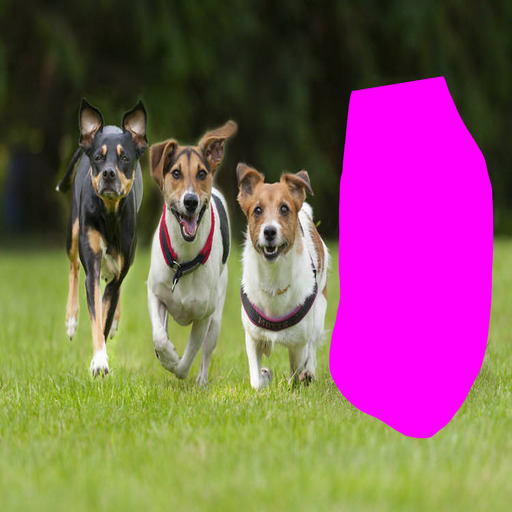}&
\includegraphics[width=.23\textwidth, height=.15\textwidth]{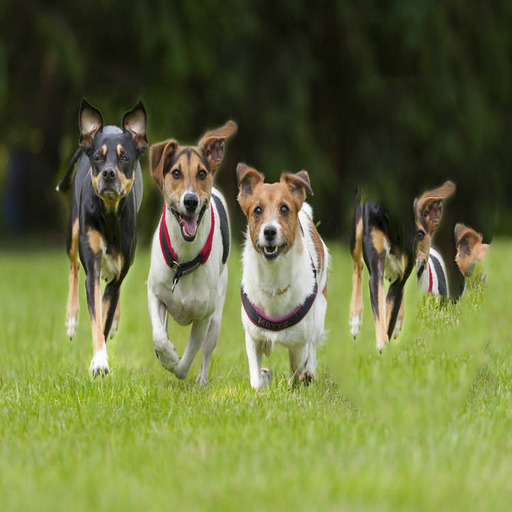}&
\includegraphics[width=.23\textwidth, height=.15\textwidth]{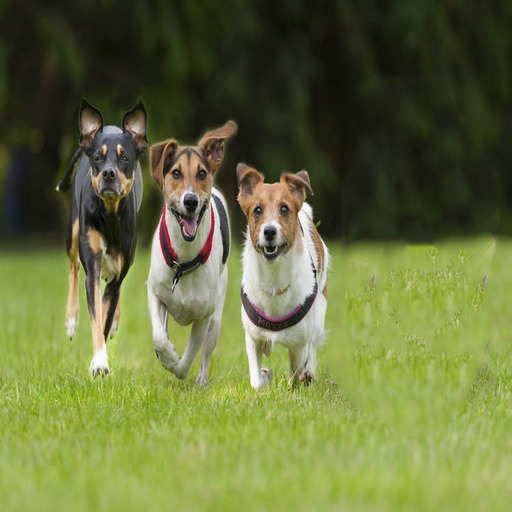} \\
  
\end{tabular}
\caption{Arbitrary object removal. From left to right: input image, object mask, Content-Aware Fill result, our result.}

 \label{fig:arbishape}
\end{figure*}

\noindent\textbf{Real-World Distractor Removal Scenario}
Finally, our algorithm is easily extended to handle arbitrary shape of holes. 
We first use a bounding rectangle to cover the arbitrary hole, which is again filled with mean-pixel values. After proper cropping and padding such that the rectangle is positioned at the center, the image is given as input to the content network. In the joint optimization, the content constraint is initialized with the output of the content network inside the arbitrary hole. The texture constraint is based on the region outside the hole. Fig.~\ref{fig:arbishape} shows several examples and its comparison with Content-Aware Fill algorithm (note that Context Encoder is unable to handle arbitrary holes explicitly so we do not compare with it here). 

\section{Conclusion}
We have advanced the state of the art in semantic inpainting using neural patch synthesis. The insight is that the texture network is very powerful in generating high-frequency details while the content network gives strong prior about the semantics and global structure. This may be potentially useful to other applications such as denoising, superresolution, retargeting and view/time interpolation. There are cases when our approach introduces discontinuity and artifacts (Fig.~\ref{fig:failure}) when the scene is complicated. In addition, the speed remains a bottleneck of our algorithm. We aim to address these issues in future work.

\section{Acknowledgment}
This research is supported in part by Adobe, Oculus \& Facebook, Huawei, the Google Faculty Research Award, the Okawa Foundation Research Grant, the Office of Naval Research (ONR) / U.S. Navy, under award number N00014-15-1-2639, the Office of the Director of National Intelligence (ODNI) and Intelligence Advanced Research Projects Activity (IARPA), under contract number 2014-14071600010, and the U.S. Army Research Laboratory (ARL) under contract W911NF-14-D-0005. The views and conclusions contained herein are those of the authors and should not be interpreted as necessarily representing the official policies or endorsements, either expressed or implied, of ODNI, IARPA, ARL, or the U.S. Government. The U.S. Government is authorized to reproduce and distribute reprints for Governmental purpose notwithstanding any copyright annotation thereon.

{\small
\bibliographystyle{ieee}
\bibliography{egbib}
}

\end{document}